\documentclass{article}

\PassOptionsToPackage{numbers, compress}{natbib}

\usepackage[final]{nips_2017}

\usepackage[T1]{fontenc}    
\usepackage[utf8]{inputenc} 
\usepackage{amsfonts}       
\usepackage{booktabs}       
\usepackage{nohyperref}     
\usepackage{microtype}      
\usepackage{nicefrac}       
\usepackage{url}            

\usepackage[labelfont=bf]{caption}
\usepackage[mathscr]{eucal}
\usepackage{algpseudocode}
\usepackage[subrefformat=parens]{subcaption}
\usepackage[usenames,dvipsnames,svgnames]{xcolor}

\usepackage{algorithmicx}
\usepackage{algorithm}
\usepackage{amsfonts}
\usepackage{amsmath}
\usepackage{amssymb}
\usepackage{amsthm}
\usepackage{bm}
\usepackage{enumitem}
\usepackage{float}
\usepackage{graphicx}
\usepackage{lipsum}
\usepackage{listings}
\usepackage{placeins}
\usepackage{textcomp}

\usepackage{tikz}
\usetikzlibrary{positioning}
\usetikzlibrary{calc}

\newtheorem{theorem}{Theorem}

\newcommand{\prog}{\mathcal{P}}
\newcommand{\kl}{\mbox{D}_{\mbox{\scriptsize KL}}}
\newcommand{\E}{\mathbb{E}}

\newcommand{\todo}[1]{}
\renewcommand{\todo}[1]{{\color{red} TODO: {#1}}}

\algnewcommand{\LeftComment}[1]{\(\triangleright\) #1}
\algnewcommand{\LineComment}[1]{\State \(\triangleright\) #1}

\newcommand{\iid}[0]{\stackrel{\mbox{\tiny i.i.d.}}{\sim}}
\newcommand{\cas}[0]{\stackrel{\mbox{\tiny a.s.}}{\to}}
\newcommand{\trace}[0]{\tau}
\newcommand{\otrace}[0]{z}
\newcommand{\orestrict}[0]{\tau|_O}

\title{Using probabilistic programs as proposals}

\author{
  Marco F.~Cusumano-Towner\\
  \small Probabilistic Computing Project\\
  \small Massachusetts Institute of Technology\\
  \texttt{marcoct@mit.edu} \\
  \And
  Vikash K.~Mansinghka\\
  \small Probabilistic Computing Project\\
  \small Massachusetts Institute of Technology\\
  \texttt{vkm@mit.edu} \\
}

\begin{document}
\tikzset{
    old inner xsep/.estore in=\oldinnerxsep,
    old inner ysep/.estore in=\oldinnerysep,
    double circle/.style 2 args={
        circle,
        old inner xsep=\pgfkeysvalueof{/pgf/inner xsep},
        old inner ysep=\pgfkeysvalueof{/pgf/inner ysep},
        /pgf/inner xsep=\oldinnerxsep+#1,
        /pgf/inner ysep=\oldinnerysep+#1,
        alias=sourcenode,
        append after command={
        let     \p1 = (sourcenode.center),
                \p2 = (sourcenode.east),
                \n1 = {\x2-\x1-#1-0.5*\pgflinewidth}
        in
            node [inner sep=0pt, draw, circle, minimum width=2*\n1,at=(\p1),#2] {}
        }
    },
    double circle/.default={2pt}{black}
}

\maketitle

\begin{abstract}
Monte Carlo inference has asymptotic guarantees, but can be slow when using generic proposals.
Handcrafted proposals that rely on user knowledge about the posterior distribution can be efficient, but are difficult to derive and implement.
This paper proposes to let users express their posterior knowledge in the form of \emph{proposal programs}, which are samplers written in probabilistic programming languages.
One strategy for writing good proposal programs is to combine domain-specific heuristic algorithms with neural network models.
The heuristics identify high probability regions, and the neural networks model the posterior uncertainty around the outputs of the algorithm.
Proposal programs can be used as proposal distributions in importance sampling and Metropolis-Hastings samplers without sacrificing asymptotic consistency, and can be optimized offline using inference compilation.
Support for optimizing and using proposal programs is easily implemented in a sampling-based probabilistic programming runtime.
The paper illustrates the proposed technique with a proposal program that combines RANSAC and neural networks to accelerate inference in a Bayesian linear regression with outliers model.
\end{abstract}

\section{Introduction}
Monte Carlo approaches to approximate inference include importance sampling, Markov chain Monte Carlo, and sequential Monte Carlo \cite{del2006sequential}.
It is easy to construct Monte Carlo inference algorithms that have asymptotic guarantees, but constructing Monte Carlo inference algorithms that are accurate in practice often requires knowledge of the posterior distribution.
Recent work in amortized inference \cite{stuhlmuller2013learning,paige2016inference,ritchie2016neurally,le2016inference} aims to acquire this knowledge by training neural networks on offline problem instances and generalizing to problem instances encountered at run-time.
We propose to use probabilistic programming languages as a medium for users to encode their domain-specific knowledge about the posterior, and to learn parameters of the program offline using an amortized inference approach to fill in gaps in this knowledge.
In our setting, the model in which inference is being performed may or may not be itself represented by a probabilistic program.

This paper makes three contributions:
First, we introduce \emph{proposal programs}, which are probabilistic programs that represent proposal distributions.
Proposal programs can use internal random choices, and report estimates of their proposal probabilities instead of true proposal probabilities.
We show that proposal programs can be used in place of regular proposal distributions within importance sampling and Metropolis-Hastings algorithms without sacrificing asymptotic consistency of these algorithms.
Second, we give a stochastic gradient descent algorithm for offline learning of proposal program parameters.
Third, we propose a class of proposal programs based on combination of domain-specific heuristic randomized algorithms for identifying high-probability regions of the target distribution with neural networks that model the uncertainty in the target distribution around the outputs of the algorithm.
We also discuss how to implement proposal programs in a sampling-based probabilistic programming runtime, and give example proposal programs written in Gen.jl \cite{cusumano2017gen}, a probabilistic language embedded in Julia \cite{bezanson2017julia}.

\section{Notation}

Consider a target distribution $\pi(z)$ on latent variable(s) $z$, and let $\tilde{\pi}(z) := c \pi(z)$ be the unnormalized target distribution where $c > 0$ is a normalizing constant.
Let $x$ denote the problem instance---this may include the observed data, or any other context relevant to defining the inference problem.
In importance sampling, we sample many values $z$ from a proposal distribution $p(z;x)$ that is parameterized by $x$, and evaluate the importance weight $\tilde{\pi}(z) / p(z;x)$ for each sample, producing a weighted collection of samples that can be used to estimate expectations under $\pi$.
In Markov chain Monte Carlo, we use one or more proposal distribution(s) $p_i(z;x)$ to construct a Markov chain that converges asymptotically to $\pi(z)$, and use iterates of one or more such chains to estimate expectations under $\pi$.
Proposal distributions may be combined in cycles or mixtures \cite{tierney1994markov}, and each proposal distributions may only update a subset of the variables in $z$, leaving the remaining variables fixed.

\section{Monte Carlo inference with proposal programs}
A \emph{proposal program} is a probabilistic program $\prog$ that takes input $x$ and makes random choices, where each random choice is given an address $i \in \mathcal{A}$, where $\mathcal{A}$ is some countable set, and where some subset $O \subseteq \mathcal{A}$ of the addresses are the \emph{output choices}.
Random choices that are not output are choices are called \emph{internal choices}.
We assume that all output choices are realized in all possible executions of the program.
We assume that all random choices are discrete---a rigorous treatment of continuous random choices is left for future work.
A \emph{trace} of a probabilistic program is a mapping $\trace : A \to V$ from the address of a random choices to its value where $A \subseteq \mathcal{A}$, and where $V$ is the set of all possible values for a random choice.
A function $z : O \to V$ from output addresses $O$ to values is called an \emph{output trace}.
Let $\orestrict : O \to V$ denote the output trace constructed by restricting a trace $\tau$ to addresses $O$.
Let $\tau \cong_O z$ if $\tau(i) = z(i)$ for all $i \in O$ (i.e. if the trace $\tau$ agrees with an output trace $z$ on the values of the output choices).
Let $p(\trace(i); \trace_{<i}, x)$ denote the probability of choice with address $i$ taking value $\trace(i)$ given the state of the program at that point in its execution as determined by the values $\trace_{<i}$ of previous random choices.
The probability of a trace $\trace$ on input $x$ is defined as:
\begin{equation}
    p(\trace; x) := \prod_{i \in A} p(\trace(i); \trace_{<i}, x)
\end{equation}
We assume that it is possible to execute the program $\prog$, possibly with some output random choices fixed to given values, and to compute the probabilities $p(\trace(i); \trace_{<i}, x)$ for output random choices $i \in O$.
Let $\trace \sim p(\cdot; x)$ denote a trace generated by executing the program on input $x$ and let $\trace \sim p(\cdot; x, \otrace)$ denote a trace generated by sampling the program on input $x$, but where output choices $i \in O$ are fixed to values given in an output trace $\otrace$ instead of being sampled ($\trace(i) = \otrace(i)$).
Note that internal random choices can occur before or after output choices in program execution, and internal random choices may depend on the value of output random choices.
We require that for all $x$ and $z$ the program satisfies $p(\tau; x, z) > 0 \implies p(\tau; x) > 0$.
That is, any trace that can be generated by executing the program with output choices fixed to $z$ must also be possible under standard program execution.
Let the product of probabilities of each output choice in a trace given the previous choices be denoted:
$p_O(\trace; x) := \prod_{i \in O} p(\trace(i); \trace_{<i}, x)$ and similarly for the internal choices $p_I(\trace; x) := \prod_{i \in A \setminus O} p(\trace(i); \trace_{<i}, x)$.
The marginal probability of an output trace $z$ is called the \emph{proposal probability}, and is given by:
\begin{equation} \label{eq:marginal}
    p(z; x) := \sum_{\trace : \trace \cong_O z} p(\trace; x)
\end{equation}

\begin{figure}[h]
\centering
\begin{tabular}{ll}
Probabilistic program & $\prog$\\
All addresses of possible random choices & $\mathcal{A}$\\
Addresses of realized random choices in an execution & $A \subseteq \mathcal{A}$\\
Possible values for random choices & $V$\\
Trace (record of random choices in an execution) & $\trace : A \to V$\\
Output random choices & $O \subseteq A$\\
Inputs to program & $x$\\
Output trace & $\otrace : O \to V$\\
Output trace generated by restricting a trace $\tau$ to outputs & $\orestrict : O \to V$\\
Sampling a trace by executing program on input $x$ & $\trace \sim p(\cdot; x)$\\
Sampling a trace with constrained output trace & $\trace \sim p(\cdot; x, \otrace)$\\
Proposal probability & $p(z; x)$
\end{tabular}
\caption{Notation for proposal programs}
\label{fig:notation}
\end{figure}

\begin{figure}[h]
\centering
\begin{subfigure}[t]{0.49\textwidth}
\begin{algorithmic}
    \Procedure{simulate}{$\prog$, $x$}
        \State $\trace \sim p(\cdot; x)$ \LeftComment{Execute $\prog$}
        \State $\otrace \gets \trace|_O$ \LeftComment{Extract output trace}
        \State $\xi \gets p(z; x)$ \LeftComment{Compute proposal probability}
        \State \Return $\left(z, \xi\right)$
    \EndProcedure
\end{algorithmic}
\end{subfigure}%
\begin{subfigure}[t]{0.49\textwidth}
\begin{algorithmic}
    \Procedure{assess}{$\prog$, $x$, $z$}
        \State $\xi \gets p(z; x)$ \LeftComment{Compute proposal probability}
        \State \Return $\xi$
    \EndProcedure
\end{algorithmic}
\end{subfigure}

\begin{subfigure}[t]{1\textwidth}
\caption{Idealized proposal program interface}
\label{fig:interface-a}
\end{subfigure}
\vspace{5mm}

\begin{subfigure}[t]{0.49\textwidth}
\begin{algorithmic}
    \Procedure{simulate}{$\prog$, $x$, $K$}
        \State $k \sim \mbox{Uniform}(1, \ldots, K)$
        \State $\trace_k \sim p(\cdot; x)$ \LeftComment{Execute $\prog$}
        \State $z \gets {\trace_k}|_O$ \LeftComment{Extract output trace}
        \For{$i \in \{1\ldots K\}\setminus \{k\}$}
            \LineComment{Execute $\prog$ with output choices fixed to $z$}
            \State $\trace_i \sim p(\cdot; x, z)$
        \EndFor
        \LineComment{Estimate proposal probability}
        \State $\hat{\xi} \gets \frac{1}{K} \sum_{i=1}^K p_O(\tau_i; x)$
        \State \Return $\left(z, \hat{\xi}\right)$
    \EndProcedure
\end{algorithmic}
\end{subfigure}%
\begin{subfigure}[t]{0.49\textwidth}
\begin{algorithmic}
    \Procedure{assess}{$\prog$, $x$, $z$, $K$}
        \For{$i \gets 1\ldots K$}
            \LineComment{Execute $\prog$ with output choices fixed to $z$}
            \State $\trace_i \sim p(\cdot; x, z)$
        \EndFor
        \LineComment{Estimate proposal probability}
        \State $\hat{\xi} \gets \frac{1}{K} \sum_{i=1}^K p_O(\tau_i; x)$
        \State \Return $\hat{\xi}$
    \EndProcedure
\end{algorithmic}
\end{subfigure}

\begin{subfigure}[t]{1\textwidth}
\caption{Approximate proposal program interface}
\label{fig:interface-b}
\end{subfigure}
\caption{
An interface that is sufficient for using proposal programs $\prog$ in importance sampling (IS) and Metropolis-Hastings (MH).
(a) shows an ideal implementation of the interface that is intractable in the general case.
(b) shows an approximate implementation of the interface that returns estimates in place of proposal probabilities.
Theorem~\ref{thm:is} and Theorem~\ref{thm:mh} show that IS and MH remain asymptotically consistent even with the approximate implementation in (b).
}
\end{figure}

In order to use the marginal distribution $p(z; x)$ on outputs of a proposal program as a proposal distribution within a standard importance sampling (IS) algorithm for a target distribution $\pi(z)$, we require the ability to sample from the proposal distribution and evaluate the proposal probability $p(z;x)$.
The proposal probability is needed to compute the importance weight $\tilde{\pi}(z) / p(z; x)$.
To use the proposal program inside standard Metropolis-Hastings (MH), we require the ability to sample $z'$ from the proposal and evaluate the forward proposal probability $p(z';x)$ for the resulting sample (where $x$ may contain the previous iterate $z_{t-1}$), as well as the ability to compute the reverse proposal probability $p(z_{t-1};x')$ (the probability that the proposal generates the reverse move, where $x'$ may contain the proposed iterate $z'$).
The forward and reverse proposal probabilities are needed to compute the acceptance probability:
\begin{equation}
\min\left\{1, \frac{\tilde{\pi}(z') p(z_{t-1}; x')}{\tilde{\pi}(z_{t-1}) p(z'; x)} \right\}
\end{equation}
Figure~\ref{fig:interface-a} shows an interface for proposal programs that, together with the ability to evaluate $\tilde{\pi}(z)$, is sufficient for implementing IS and MH.
The \textproc{simulate} procedure executes the proposal program, and returns the output trace $z$ and the proposal probability $p(z; x)$.
The \textproc{assess} procedure takes an output trace $z$ and returns $p(z; x)$.
However, evaluating $p(z;x)$ is intractable in the general case because Equation~(\ref{eq:marginal}) contains a number of terms that is exponential in the number of internal random choices.
Therefore, we propose an approximate implementation of the interface, shown in Figure~\ref{fig:interface-b}.
The approximate \textproc{simulate} procedure executes the proposal program, and returns the output trace $z$ and a specific estimate $\hat{\xi}$ of the proposal probability $p(z; x)$.
The approximate \textproc{assess} procedure takes an output trace $z$ and returns a specific estimate $\hat{\xi}$ of $p(z; x)$.
Algorithm~\ref{alg:is} shows an IS algorithm that uses the approximate \textproc{simulate} procedure of Figure~\ref{fig:interface-b}, and Algorithm~\ref{alg:mh} shows a MH transition operator that uses the approximate \textproc{simulate} and \textproc{assess} procedures.
The remainder of the section shows that substituting the approximate procedures in Figure~\ref{fig:interface-b} in place of the idealized procedures in Figure~\ref{fig:interface-a} within IS and MH preserves the asymptotic consistency of both algorithms.

\begin{algorithm}
\begin{algorithmic}
\Require Target distribution $\pi(z)$; unnormalized target distribution $\tilde{\pi}(z) = c \pi(z)$ for some $c>0$; test function $f(z)$; proposal program $\prog$; proposal arguments $x$; integers $N,K \ge 1$.
\For{$i \gets 1, \ldots, N$}
    \State $(z^{(i)}, \hat{\xi}^{(i)}) \sim \textproc{simulate}(\prog, x, K)$
    \State $w^{(i)} \gets \tilde{\pi}(z^{(i)}) / \hat{\xi}^{(i)}$
\EndFor
\State $\hat{\mu}_N \gets \displaystyle \frac{\sum_{i=1}^N w^{(i)} f(z^{(i)})}{\sum_{i=1}^N w^{(i)}}$
\State \Return $\hat{\mu}_N$
\end{algorithmic}
\caption{Importance sampling using a proposal program}
\label{alg:is}
\end{algorithm}

\begin{theorem} \label{thm:is}
If $\pi(z)$ is a target distribution, $\prog$ is a proposal program such that $\pi(z) > 0 \implies p(z; x) > 0$ for all $x$, $f$ is a function where $\mathbb{E}_{z \sim \pi}[f(z)] = \mu < \infty$, then $\hat{\mu}_N \cas \mu$ where $\hat{\mu}_N$ is the importance sampling estimate returned by Algorithm~\ref{alg:is} using the approximate \textproc{simulate} procedure of Figure~\ref{fig:interface-b}.
\end{theorem}
See Appendix~\ref{sec:is-proof} for the proof, which is based on an auxiliary variable construction.
Algorithm~\ref{alg:mh} shows a Metropolis-Hastings transition operator that uses the \textproc{simulate} and \textproc{assess} procedures of Figure~\ref{fig:interface-b}.
\begin{algorithm}
\begin{algorithmic}
\Require Unnormalized distribution $\tilde{\pi}(z)$; proposal program $\prog$; integer $K \ge 1$; previous iterate $z_{t-1}$.
\State $(z', \hat{\xi}') \sim \textproc{simulate}(\prog, z_{t-1}, K)$
\State $\hat{\xi}_{t-1} \sim \textproc{assess}(\prog, z_{t-1}, z', K)$
\State $\alpha \gets \displaystyle \frac{\tilde{\pi}(z') \hat{\xi}_{t-1}}{\tilde{\pi}(z_{t-1}) \hat{\xi}'}$
\State $r \sim \mbox{Uniform}(0, 1)$
\If{$r \le \alpha$}
    \State $z_t \gets z'$
\Else
    \State $z_t \gets z_{t-1}$
\EndIf
\end{algorithmic}
\caption{Metropolis-Hastings transition using a proposal program}
\label{alg:mh}
\end{algorithm}

\begin{theorem} \label{thm:mh}
The transition operator of Algorithm~\ref{alg:mh} using the approximate \textproc{simulate} and \textproc{assess} procedures of Figure~\ref{fig:interface-b} admits $\pi(x)$ as a stationary distribution.
\end{theorem}
See Appendix~\ref{sec:mh-proof} for the proof.
Note that this transition operator may be composed with other transition operators to construct an ergodic Markov chain.
The non-asymptotic accuracies of Algorithm~\ref{alg:is} and Algorithm~\ref{alg:mh} depend on the choice of $K$, the number of executions of the program used within each invocation of \textproc{simulate} or \textproc{assess}.
As $K \to \infty$, the algorithms behave like the corresponding IS and MH algorithms using the exact proposal probabilities.
For finite $K=1$, the deviation of these algorithms from the $K \to \infty$ limit depends on the details of the proposal program.
If $z$ is independent of $y$ or if $y$ is deterministic, $K=1$ provides exact proposal probabilities.
We expect that small $K$ may be sufficient for proposal programs which have low mutual information between the internal trace and the output trace.
A detailed analysis of how performance depends on $K$ is left for future work.

The procedures in Figure~\ref{fig:interface-b} can be easily implemented in a sampling-based probabilistic programming runtime that records a probability for each random choice in a trace or `random database' \cite{wingate2011lightweight}.
We implemented such a runtime for Gen.jl \cite{cusumano2017gen}, a probabilistic programming language embedded in Julia \cite{bezanson2017julia}.
In Gen.jl, a probabilistic program is a Julia function that has been annotated with the \texttt{@probabilistic} keyword, and in which some random choices have been annotated with unique addresses using the \texttt{@choice} keyword.

\section{Learning proposal program parameters}
This section describes a technique to learn the parameters of a proposal program during an offline `inference compilation' phase, prior to use in importance sampling or Metropolis-Hastings.
Let $\pi(z; x)$ denote a family of target distributions indexed by the problem instance $x$.
We consider a proposal program $\prog$ parameterized by $\theta$, with distribution on traces denoted $p(\trace; x, \theta)$.
We assume that it is possible to sample from a training distribution on the arguments $x$ of the proposal program and the outputs trace $z$ of the proposal program, denoted $r(x, z)$, where $r(z|x) > 0 \iff \pi(z;x) > 0$.
We factor the training distribution into a distribution on arguments $r(x)$, and a desired distribution on outputs given arguments $r(z|x)$.
If the model $\pi$ in which we are doing inference is a generative model where $\pi(z, x)$ is a joint distribution on latent variables $z$ and data (observation) $x$, then we can define $\pi(z; x) := \pi(z|x)$, and $r(x, z) := \pi(z, x)$, so that the argument distribution is $\pi(x)$ and the desired output distribution is the posterior $\pi(z|x)$.
For this choice of $r$, sampling from the training distribution is achieved by ancestral sampling of the latents and observations from the generative model.
A different training distribution based on gold-standard SMC sampling instead of joint generative simulation was previously used in \cite{ritchie2016neurally}.
Given $r(x, z)$, we seek to find $\theta$ that solve the following optimization problem, where $\kl$ denotes the Kullback-Leibler (KL) divergence:
\begin{align}
    \min_{\theta} \E_{x \sim r(\cdot)} \left[ \kl(r(z|x) || p(z; x, \theta)) \right]
\end{align}
This is equivalent to the maximizing the expected conditional log likelihood of the training data:
\begin{align} \label{eq:compilation-objective}
    \max_{\theta} J(\theta) = \max_{\theta} \E_{z, x \sim r(\cdot, \cdot)} \left[ \log p(z; x, \theta) \right]
\end{align}
Because the proposal program may use internal random choices, it may not be possible to evaluate $\log p(z; x, \theta)$, making direct optimization of $J(\theta)$ difficult.
Therefore, we instead maximize a lower bound $J^K(\theta)$ on $J(\theta)$:
\begin{align}
    J^K(\theta) := \E_{x, z \sim r(\cdot, \cdot)} \left[ \E_{\trace_{1:K} \iid p(\cdot; x, \theta, z)} \left[ \log \hat{\xi}(\trace_{1:K}, x, \theta) \right] \right] \le J(\theta)
\end{align}
where $\hat{\xi}(\trace_{1:K}, x, \theta) := \frac{1}{K} \sum_{k=1}^K p_O(\trace_k; x, \theta)$.
We maximize $J^K(\theta)$ using stochastic gradient ascent.
For some $j \in \{1, \ldots, K\}$, let $\hat{\xi}(\trace_{-j}, x, \theta) := \frac{1}{K-1} \sum_{k\ne j}^K p_O(\trace_k; x, \theta)$.
Let $W_k := p_O(\trace_k; x, \theta)/(\sum_{j=1}^K p_O(\trace_j; x, \theta))$.
Also, define functions $h$ and $g$:
\begin{align}
    g(x, \trace_{1:K}, \theta) &:= \sum_{k=1}^K \left( \log \hat{\xi}(\trace_{1:K}, x, \theta) - \log \hat{\xi}(\trace_{-k}, x, \theta)) \right) \nabla_{\theta} \log p_I(\trace_k; x, \theta)\\
    h(x, \trace_{1:K}, \theta, W_{1:K}) &:= \sum_{k=1}^K W_k \nabla_{\theta} \log \hat{\xi}(\trace_{1:K}, x, \theta)
\end{align}
We then use the following unbiased estimator of the gradient $\nabla_{\theta} J^K(\theta)$ based on the `per-sample' baseline of \cite{mnih2016variational}: 
\begin{align}
    g(x, \trace_{1:K}, \theta) + h(x, \trace_{1:K}, \theta) \;\;\mbox{for}\;\; x, z \sim r(\cdot, \cdot) \;\;\mbox{and}\;\; \trace_{1:K} \iid p(\cdot; x, \theta, z)\nonumber
\end{align}
The first term accounts for the effect of $\theta$ on the internal random choices and the second term accounts for the direct effect of $\theta$ on the output random choices.
The resulting algorithm is shown below:
\begin{algorithm}
\begin{algorithmic}
\Require Training distribution $r(x, z)$; proposal program $\prog$; integer $K \ge 1$, initial parameters $\theta_0$, mini-batch size $M$.
\State $t \gets 0$
\While{not converged}
    \State $t \gets t + 1$
    \For{$m \gets 1, \ldots, M$}
        \State $x^{(m)}, z^{(m)} \sim r(\cdot, \cdot)$ \Comment{Sample from training distribution}
        \For{$k \gets 1, \ldots, K$}
            \State $\trace_k^{(m)} \sim p(\cdot; x^{(m)}, \theta_{t-1}, z^{(m)})$ \Comment{Execute $\prog$ with output choices fixed to $z^{(m)}$}
        \EndFor
        \State $\Delta \theta^{(m)} \gets g(x^{(m)}, \trace^{(m)}_{1:K}, \theta_{t-1}) + h(x^{(m)}, \trace^{(m)}_{1:K}, \theta_{t-1})$
    \EndFor
    \State $\theta_t \gets \theta_{t-1} + \rho_t \frac{1}{M} \sum_{m=1}^M \Delta \theta^{(m)}$
\EndWhile
\State \Return $\theta_t$
\end{algorithmic}
\caption{Offline optimization of proposal program}
\label{alg:optimization}
\end{algorithm}

Algorithm~\ref{alg:optimization} can be easily implemented on top of a sampling-based probabilistic programming runtime that includes reverse-mode automatic differentiation.
The runtime must provide (1) the log probability of output random choices $\log p_O(\trace; x, \theta)$ and its gradient $\nabla_{\theta} \log p_O(\trace; x, \theta)$ and (2) the gradient of the log probability of internal random choices, $\nabla_{\theta} \log p_I(\trace; x, \theta)$.
The gradient $\nabla_{\theta} \log \hat{\xi}(\trace_{1:K}, x, \theta)$ appearing in $h(x, \trace_{1:K}, \theta)$ can be computed from the collection of $\nabla_{\theta} \log p_O(\trace_k; x, \theta)$ for each $k$.

Note that the log probability of internal random choices $\log p_I(\trace; x, \theta)$ is not required (but its gradient is).
Recall that the log probability of the internal random choices is the sum of the log probability of each internal random choice in the internal trace.
For internal random choices that do not depend \emph{deterministically} on $\theta$, the log probability contribution does not depend on $\theta$.
Therefore, it suffices for the runtime to only instrument the internal random choices that deterministically depend on $\theta$.
This permits use of black box code within the proposal program, which can reduce the overhead introduced by the probabilistic runtime in performance-critical sections of the proposal, and is important for our proposed methodology for using proposal programs in Section~\ref{sec:methodology}.
We implemented Algorithm~\ref{alg:optimization} on top of the Gen.jl runtime \cite{cusumano2017gen}.
In Gen.jl, the user annotates random choices in a probabilistic program with a unique identifier.
The user is free to use randomness in a probabilistic program without annotating it.
In our implementation of Algorithm~\ref{alg:optimization} we only accumulate the gradient $\nabla_{\theta} \log p_I(\trace; x, \theta)$ taking into account random choices that were annotated, relying on the user to ensure that there is no deterministic dependency of un-annotated random choices on $\theta$.
In future work, automatic dependency analysis based on the approach of \cite{schulman2015gradient} could remove this reliance on the user.

\section{Application: Proposals based on randomized mode-finding algorithms} \label{sec:methodology}
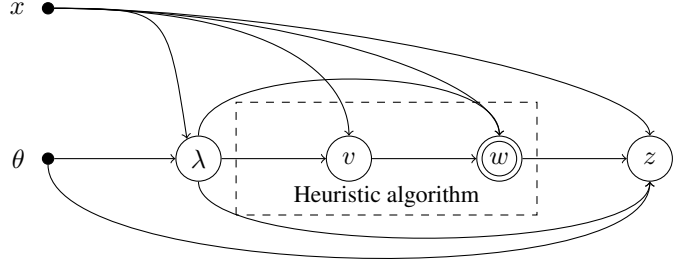
\begin{figure}
\setlength{\abovecaptionskip}{10pt}
\centering
\begin{tikzpicture}
\tikzstyle{line} = [draw, thick]
\node[fill,minimum size=0.15cm,inner sep=0pt] (x) at (0, 2) [circle, draw] {};
\node[left=0.1cm of x] (xlabel) {$x$};
\node[fill,minimum size=0.15cm,inner sep=0pt] (theta) at (0, 0) [circle, draw] {};
\node[left=0.1cm of theta] (thetalabel) {$\theta$};
\node[minimum size=0.6cm,inner sep=0pt] (lambda) at (2, 0) [circle, draw] {$\lambda$};
\node[minimum size=0.6cm,inner sep=0pt] (v) at (4, 0) [circle, draw] {$v$};
\node[minimum size=0.6cm,inner sep=0pt,double circle] (state) at (6, 0) [circle, draw] {$w$};
\node[minimum size=0.6cm,inner sep=0pt] (z) at (8, 0) [circle, draw] {$z$};
\draw [->] (theta) -- (lambda);
\draw [->] (lambda) -- (v);
\draw [->] (v) -- (state);
\draw [->] (state) -- (z);
\draw [->] (node cs:name=lambda, anchor=north)
           .. controls +(0,1) and +(0,1)
           .. (node cs:name=state, anchor=north);
\draw [->] (node cs:name=x, anchor=east)
           .. controls +(1.5,0)
           .. (node cs:name=lambda, anchor=120);
\draw [->] (node cs:name=x, anchor=east)
           .. controls +(3,0) and +(0,0.75)
           .. (node cs:name=v, anchor=north);
\draw [->] (node cs:name=x, anchor=east)
           .. controls +(3,0) and +(0,0.75)
           .. (node cs:name=state, anchor=north);
\draw [->] (node cs:name=x, anchor=east)
           .. controls +(3,0) and +(0,0.75)
           .. (node cs:name=z, anchor=north);
\draw [->] (node cs:name=theta, anchor=south)
           .. controls +(0,-1.5) and +(0,-1.5)
           .. (node cs:name=z, anchor=south);
\draw [->] (node cs:name=lambda, anchor=south)
           .. controls +(0,-1) and +(0,-1)
           .. (node cs:name=z, anchor=south);
\draw [dashed] (2.5,-0.75) rectangle (6.5,0.75);
\node[] (boxtext) at (4.5, -0.5) {\small Heuristic algorithm};
\end{tikzpicture}
\caption{
Proposed high-level data flow for a proposal program based on a heuristic randomized algorithm.
$x$ is the input to the program, $\theta$ are the parameters of the program to be optimized, $\lambda$ are the parameters of the heuristic algorithm that are difficult to set manually, and $z$ is the output of the proposal program.
The internal random choices $u$ of the proposal program include $\lambda$ and any internal random choices $v$ made by the heuristic algorithm, which need not be instrumented by the probabilistic runtime.
}
\label{fig:heuristic}
\end{figure}
Because Algorithm~\ref{alg:optimization} minimizes the KL divergence from the posterior (or its proxy) to the proposal distribution, optimization will tend to force the proposal to have sufficient variability to cover the posterior mass, which provides a degree of robustness even when the heuristic is brittle.
On problem instances where the heuristic gives wrong answers, the proposal variability should increase.
On problem instances where the heuristic produces values near the posterior modes, the proposal variability should attempt to match the variability of each posterior mode.

Proposal programs permit the inference programmer to express their knowledge about the target distribution in the form of a sampling program.
Consider the case when the user possesses a heuristic randomized algorithm that finds areas of high probability of the target distribution.
Although such an algorithm contains some knowledge about the target distribution, the algorithm in isolation is not useful for performing sound probabilistic inference---it does not necessarily attempt to represent the variability with the mode(s) of the target distribution, and its support may not even include the posterior support.
The heuristic may also be brittle---it have parameters that need to be tuned for it to work on a particular problem instance, or it may fail completely on some problem instances.

To construct robust and sound probabilistic inference algorithms that can take advantage of heuristic randomized algorithms, we construct proposal programs that use the heuristic algorithm as a subroutine, and use the proposal program in importance sampling (Algorithm~\ref{alg:is}) or MCMC (Algorithm~\ref{alg:mh}).
Let $x$ denote the problem instance (e.g. the observed data or context variables); and let $\lambda$, $v$, and $w$ denote the parameters, internal random choices, and output of the algorithm, respectively.
To construct the proposal program from the heuristic algorithm, we:
\begin{enumerate}
\item Make the parameters that are difficult to set manually into random choices $\lambda$ whose distribution may depend on $x$ (other parameters can be hardcoded or can have a hardcoded dependency on $x$).
\item Add random choices $z$ that are equal to the output $v$ of the heuristic algorithm plus some amount of noise, such that the support of $z$ given any $w$ and $x$ includes the support of the posterior $\pi(z)$.
The choices $z$ are the output choices of the proposal program.
\item Parametrize the distribution of $\lambda$ given $x$ and the distribution of $z$ given $x$ and $w$ using the output of neural network(s) with parameters $\theta$.
The networks take as input $x$ and generate parameter values $\lambda$ for the heuristic algorithm, as well as the amount of noise used to add to the output $w$ of the algorithm.
\end{enumerate}
Figure~\ref{fig:heuristic} shows the high-level data flow of the resulting proposal program.
We optimize the parameters $\theta$ during an offline inference compilation phase using Algorithm~\ref{alg:optimization}.
Note that the internal random choices of the heuristic algorithm ($v$) do not depend deterministically on $\theta$.
Therefore, to use Algorithm~\ref{alg:optimization}, the heuristic algorithm need not be instrumented by the probabilistic runtime, and can be fast black-box code.

\section{Example: Proposal program combining RANSAC with a neural network}

\begin{figure}[h]
\begin{lstlisting}[frame=single, numbers=left]
function ransac(xs, ys, params::RANSACParams)
    best_num_inliers::Int = -1
    best_slope::Float64 = NaN
    best_intercept::Float64 = NaN
    for i=1:params.num_iters

        # Select a random pair of points
        rand_ind = StatsBase.sample(1:length(xs), 2, replace=false)
        subset_xs = xs[rand_ind]
        subset_ys = ys[rand_ind]
        
        # Estimate slope and intercept using least squares
        A = hcat(subset_xs, ones(length(subset_xs)))
        slope, intercept = A\subset_ys
        
        # Count the number of inliers for this (slope, intercept) hypothesis
        ypred = intercept + slope * xs
        inliers = abs.(ys - ypred) .< params.epsilon
        num_inliers = sum(inliers)

        if num_inliers > best_num_inliers
            best_slope, best_intercept = slope, intercept
            best_num_inliers = num_inliers
        end
    end

    # return the hypothesis that resulted in the most inliers
    (best_slope, best_intercept)
end
\end{lstlisting}
\vspace{-2mm}
\caption{A Julia implementation of a RANSAC-based heuristic algorithm for generating hypotheses (lines parameterized by a slope and intercept) that explain a given data set of x-y coordinates.}
\label{fig:ransac}
\end{figure}

\begin{figure}[h]
\begin{lstlisting}[frame=single, numbers=left]
@probabilistic function ransac_neural_proposal(xs, ys, params)

    # Generate parameters for RANSAC using learned parameters
    epsilon = @choice(gamma(exp(params.eps_alpha),
                            exp(params.eps_beta)), "epsilon")
    num_iters = @choice(categorical(params.iter_dist), "iters")
    ransac_params = RANSACParams(num_iters, epsilon)

    # Run RANSAC (uses many un-annotated random choices)
    slope_guess, intercept_guess = ransac(xs, ys, ransac_params)

    # Predict output variability using learned neural network
    nn_hidden = ewise(sigmoid, params.h_weights * vcat(xs, ys) + params.h_biases)
    nn_out = params.out_weights * nn_hidden + params.out_biases
    slope_scale, intercept_scale = (exp(nn_out[1]), exp(nn_out[2]))

    # Add noise
    slope = @choice(cauchy(slope_guess, slope_scale), "slope")
    intercept = @choice(cauchy(intercept_guess, intercept_scale), "intercept")

    # Generate outlier statuses from conditional distribution
    for (i, (x, y)) in enumerate(zip(xs, ys))
        p_outlier = conditional_outlier(x, y, slope, intercept)
        @choice(flip(p_outlier), "outlier-$i")
    end
end
\end{lstlisting}
\vspace{-2mm}
\caption{
Proposal program in Gen.jl that uses RANSAC and a neural network to predict the line, followed by conditional sampling for the outlier variables.
The output choices $O$ are \texttt{"slope"}, \texttt{"intercept"}, and \texttt{"outlier-1"}, \texttt{"outlier-2"}, etc.
In Gen.jl, the set of output choices $O$ is defined when querying the program using the methods of Figure~\ref{fig:interface-b}, not in the program text itself.
}
\label{fig:custom}
\end{figure}

\begin{figure}[h]
\begin{lstlisting}[frame=single, numbers=left]
@probabilistic function neural_proposal(xs, ys, params)

    # Generate randomness
    dim = 2
    latent = mvnormal(zeros(dim), eye(dim))

    # Construct feature vector
    features = vcat(latent, xs, ys)

    # Use neural network to predict parameters of line distribution
    hidden = ewise(sigmoid, params.hidden_weights * features + params.hidden_biases)
    output = params.output_weights * hidden + params.output_biases
    slope_mu = output[1]
    intercept_mu = output[2]
    slope_var = exp(output[3])
    intercept_var = exp(output[4])

    # Generate line
    slope = @choice(cauchy(slope_guess, slope_scale), "slope")
    intercept = @choice(cauchy(intercept_guess, intercept_scale), "intercept")

    # Generate outlier statuses from conditional distribution
    for (i, (x, y)) in enumerate(zip(xs, ys))
        p_outlier = conditional_outlier(x, y, slope, intercept)
        @choice(flip(p_outlier), "outlier-$i")
    end
end
\end{lstlisting}
\vspace{-2mm}
\caption{Proposal program in Gen.jl that a neural network to predict the line, followed by conditional sampling for the outlier variables}
\label{fig:nn}
\end{figure}

We illustrate Algorithm~\ref{alg:is} and approach of Section~\ref{sec:methodology}, on a Bayesian linear regression with outliers inference problem.
The data set is a set of x-y pairs $\{(x_i,y_i)\}_{i=1}^N$, the latent variables $z$ consist of the parameters of the line $(\alpha_1, \alpha_2)$ and binary variables $o_i$ indicating whether each data point $i$ is an outlier or not.
The generative model $\pi(z, y)$, which is conditioned on the x-coordinates $x$, is defined below:
\begin{align}
\alpha_1 &\sim \mbox{Normal}(0, 1) \;\; \mbox{(slope)}\nonumber\\
\alpha_2 &\sim \mbox{Normal}(0, 2) \;\; \mbox{(intercept)}\nonumber\\
o_i &\sim \mbox{Bernoulli}(0.1) \; \mbox{for} \; i=1\ldots N\;\;\mbox{(outlier or inlier)}\nonumber\\
y_i | \alpha_1, \alpha_2, x_i, o_i &\sim \left\{ 
\begin{array}{ll}
\mbox{Normal}(\alpha_1 x_i + \alpha_2, 1) &: o_i = 0 \;\;\mbox{(inlier)}\\
\mbox{Normal}(\alpha_1 x_i + \alpha_2, 5.8) &: o_i = 1 \;\;\mbox{(outlier)}
\end{array}
\right.\nonumber
\end{align}
The unnormalized target distribution $\tilde{\pi}(z)$ is the joint probability $\pi(z, y)$.

Random sample consensus (RANSAC, \cite{fischler1981random}) is an iterative algorithm that can quickly find lines near the posterior modes of the posterior on this problem, using deterministic least-squares algorithm within each iteration.
Figure~\ref{fig:ransac} shows a Julia implementation of a RANSAC algorithm for our linear-regression with outliers problem.
However, the output distribution of the algorithm is an atomic set that does not sufficiently support the posterior distribution on lines, which is defined on $\mathbb{R}^2$.
Also, the algorithm contains a parameter \texttt{epsilon} that needs to be roughly calibrated to the inlier variability in a particular data set, and it is not clear how many iterations of the algorithm to run.

To make use of RANSAC for robust and sound probabilistic inference, we wrote a proposal program in Gen.jl following the methodology proposed in Section~\ref{sec:methodology}, optimized the unknown parameters of the resulting proposal programs using Algorithm~\ref{alg:optimization}, and employed the optimized proposal programs within importance sampling (Algorithm~\ref{alg:is}).
The proposal program, shown in Figure~\ref{fig:custom}, (1) generates the parameter \texttt{epsilon} and the number of iterations of the RANSAC algorithm from distributions whose parameters are part of $\theta$, then (2) runs RANSAC, and (3) adds Cauchy-distributed noise to the resulting line parameters (slope and intercept), where the variability of the noise is determined by the output of a neural network whose parameters are part of $\theta$, and finally (4) samples the outlier binary variables from their conditional distributions given the line.
Note that the random choices inside the \texttt{ransac} function are not annotated---the RANSAC procedure is executed as regular Julia code by Gen.jl's probabilistic runtime without incurring any overhead.

\begin{figure}[h]
\centering
\begin{subfigure}[t]{0.25\textwidth}
    \centering
    \includegraphics[width=\textwidth]{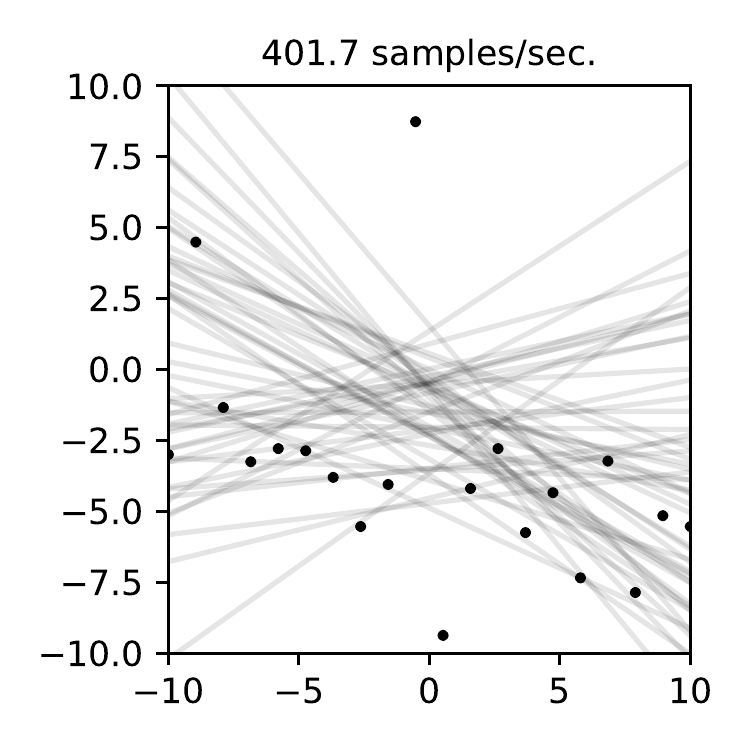}
    \caption{IS (prior)}
    \label{fig:bar}
\end{subfigure}%
\begin{subfigure}[t]{0.25\textwidth}
    \centering
    \includegraphics[width=\textwidth]{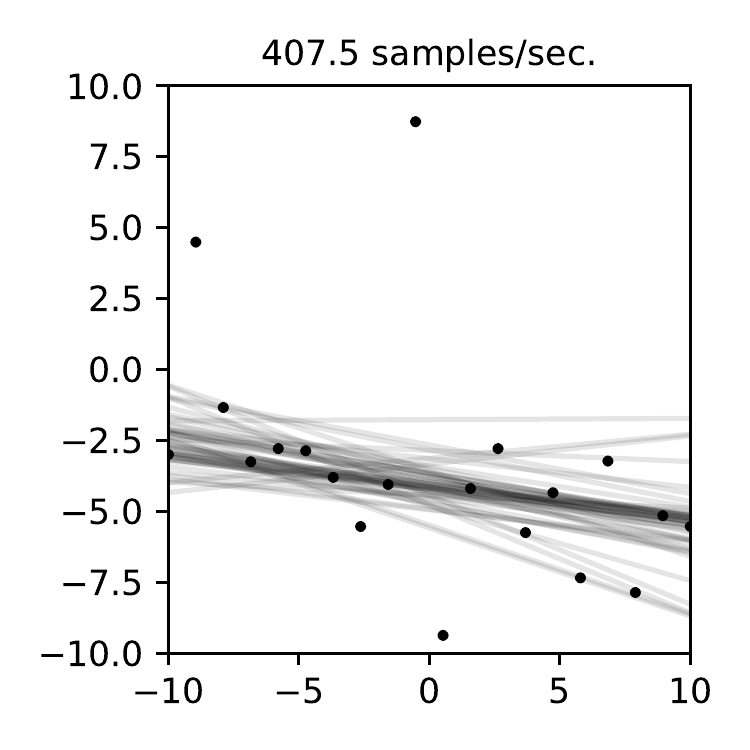}
    \caption{IS (RANSAC+NN)}
    \label{fig:qux}
\end{subfigure}%
\begin{subfigure}[t]{0.25\textwidth}
    \centering
    \includegraphics[width=\textwidth]{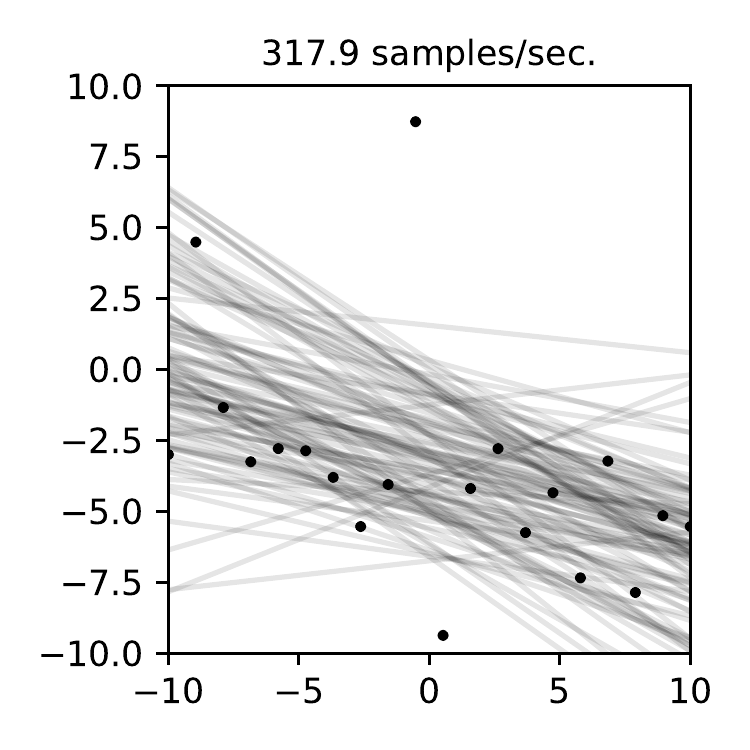}
    \caption{IS (NN)}
    \label{fig:qux}
\end{subfigure}%
\begin{subfigure}[t]{0.24\textwidth}
    \centering
    \includegraphics[width=\textwidth]{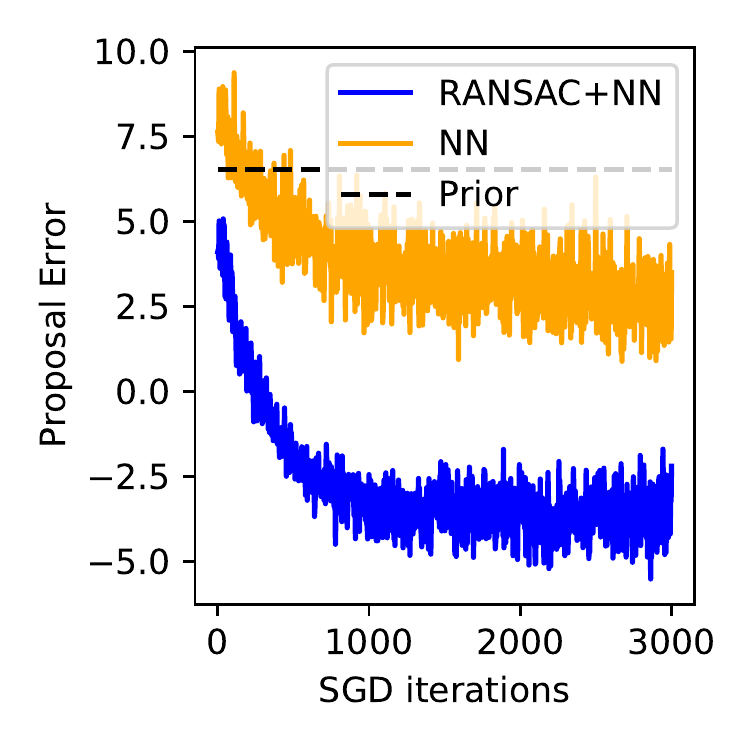}
    \caption{Training objective}
    \label{fig:foo}
\end{subfigure}
\caption{
(a) shows a dataset (points), and approximate posterior samples (lines) from an importance sampling algorithm using a prior proposal.
(b) shows samples produced by Algorithm~\ref{alg:is} using a proposal program (RANSAC+NN, Figure~\ref{fig:custom}) that combines RANSAC with a neural network.
(c) shows samples produced Algorithm~\ref{alg:is} using a proposal program (NN, Figure~\ref{fig:nn}) based on a neural network.
Six particles ($N = 6$) were used for both proposals.
Significantly more accurate samples are obtained with comparable throughput using the RANSAC+NN proposal program.
(d) shows the estimated approximation error of the three proposals as the parameters of proposals are tuned offline using ADAM.
Error is quantified using the expected KL divergence from the target distribution to the proposal distribution up to an unknown constant that does not depend on the proposal distribution, where the expectation is taken under datasets sampled from the model.
 }
\label{fig:example}
\end{figure}

For comparison, we also implemented a proposal program (Figure~\ref{fig:nn}) that does not use RANSAC, but instead generates the slope and intercept of the line from a distribution that depends on a neural network.
We optimized both proposal programs using a modified version of Algorithm~\ref{alg:optimization} that uses ADAM \cite{kingma2014adam} instead of standard SGD.
We used the generative model $\pi(z, y)$ as the training distribution, and we used $K=100$ during training and minibatches of size $8$, and $3000$ iterations.
The results are shown and discussed in Figure~\ref{fig:example}.

\section{Related work} \label{sec:related}
Some probabilistic programming systems support combining automated and custom inference strategies \cite{mansinghka2014venture}.
However, custom proposal distributions in \cite{mansinghka2014venture} require user implementation of a density-evaluation function for the proposal (mirroring the interface of Figure~\ref{fig:interface-a}).
Other probabilistic programming systems support the specification of `guide programs' for use in importance sampling, sequential Monte Carlo, or variational inference \cite{ritchie2016neurally,ritchie2016deep}.
However, guide programs follow the control flow of a model program, or are restricted to using the same set of random choices.
We focus on using general probabilistic programs to define proposal distributions, whether in the context of probabilistic programs for the model or not.

The Monte Carlo algorithms presented here derive their theoretical justification from auxiliary-variable constructions similar to those of pseudo-marginal MCMC \cite{andrieu2009} and random-weight particle filter \cite{fearnhead2008particle}.
In particular, our Metropolis-Hastings algorithm can be seen as an application of the general auxiliary-variable MCMC formalism of \cite{storvik2011flexibility} to the setting where proposals are defined as probabilistic programs.
 
Recent work in `amortized' or `compiled' inference has studied techniques for offline optimizing of proposal distributions in importance sampling or sequential Monte Carlo \cite{stuhlmuller2013learning,paige2016inference,ritchie2016neurally,le2016inference}.
Others have applied similar approaches to optimize proposal distributions for use in MCMC \cite{cusumano2017probabilistic,wang2017neural}.
However, the focus of these efforts is optimizing over neurally-parameterized distributions that inherit their structure from the generative model being targeted, do not contain their own internal random choices, and are therefore not suited to use with heuristic randomized algorithms.
In contrast, we seek to allow the user to easily express and compute with custom proposal distributions that are defined as arbitrary probabilistic programs that are independent of any possible structure in the probabilistic model being targeted, and may include `internal' random choices not present in the target model.

There has been much recent work in probabilistic machine learning on training generative latent variable models using using stochastic gradient descent \cite{kingma2013auto,mnih2014neural,mnih2016variational}.
Our procedure for optimizing proposal programs is an application of the Monte Carlo variational objective approach of \cite{mnih2016variational} to the setting where the generative model is itself a proposal distribution in a different probabilistic model.
The observation that random variables can permit optimization of probabilistic computations that utilize black-box randomized code has been previously observed and used in reinforcement learning \cite{williams1992simple,schulman2015gradient}.

\section{Discussion}
This paper formalized proposal programs, discussed how to implement proposal programs in a sampling-based probabilistic runtime, showed how to use proposal programs within importance sampling and Metropolis-Hastings, how to optimize proposal programs offline, and suggested an application of the formalism to using randomized heuristics to accelerate Monte Carlo inference.
Several directions for future work seem promising.

First, we note that the efficiency of Monte Carlo inference with proposal programs depends on the number of internal replicates $K$ used within the proposed \textproc{simulate} and \textproc{assess} procedures.
It is important to better characterize how the efficiency depends on $K$.
Also, the proposed \textproc{simulate} and \textproc{assess} procedures make use of forward execution of the proposal program to generate traces that are used for estimating the proposal probability.
Following the `probabilistic module' formalism \cite{cusumano2016encapsulating}, other distributions on traces that attempt to approximate the conditional distribution on traces given output choices can be used instead for more accurate proposal probability estimates (and therefore more efficient Monte Carlo inference for a fixed proposal program).
These distributions, which constitute nested inference samplers, should be able to make use of existing inference machinery in probabilistic programming languages that was originally developed for inference in model programs.
It also seems promising to explore proposal programs that combine neural networks and domain-specific heuristic algorithms in different ways.
For example, a proposal program may contain a switch statement that decides whether to propose using a heuristic (as in Section~\ref{sec:methodology}) or according to a pure-neural network proposal (as shown in Figure~\ref{fig:nn}), where the probability of taking the two different paths can itself be predicted from the problem instance using a neural network.
The proposal program formalism also suggests that other approaches for inference in probabilistic programs such as symbolic integration \cite{gehr2016psi} could find applications in estimating or computing proposal probabilities.
Finally, a rigorous theoretical analysis of importance sampling, Metropolis-Hastings, and offline optimization with proposal programs that contain continuous random variables would be useful.

\subsubsection*{Acknowledgments}
This research was supported by DARPA (PPAML program, contract number
FA8750-14-2-0004), IARPA (under research contract 2015-15061000003), the Office
of Naval Research (under  research  contract N000141310333), the Army Research
Office (under agreement number W911NF-13-1-0212),  and gifts from Analog
Devices and Google.
This research was conducted with Government support under and awarded by DoD, Air Force Office of Scientific Research, National Defense Science and Engineering Graduate (NDSEG) Fellowship, 32 CFR 168a.

\bibliographystyle{unsrt}
\bibliography{references} 

\appendix

\section{Proofs} \label{sec:proof}

\subsection{Consistency of importance sampling using proposal program} \label{sec:is-proof}
Consider a proposal program $\prog$ and a target distribution $\pi(z)$ such that for all $z$ where $\pi(z) > 0$ there exists a trace $\trace$ of $\prog$ where $\trace \cong_O z$ and $p(\trace; x) > 0$.
Also assume that $p(\trace; x, z) > 0 \implies p(\trace; x) > 0$ for all $\trace \cong_O z$.
Consider the following extended target distribution:
\begin{align}
    \pi(z, \trace_{1:K}) := \pi(z) \prod_{i=1}^K p(\trace_i; x, z)
\end{align}
Note that $\pi(z, \trace_{1:K})$ could hypothetically be sampled from by first sampling from $z \sim \pi(\cdot)$ and then executing $\prog$ using fixed output choices $z$.
Consider the following extended proposal distribution:
\begin{align}
    q(z, \trace_{1:K}) := \left\{ \begin{array}{ll}
        \frac{1}{K} \sum_{k=1}^K p(\trace_k; x) \prod_{i \ne k} p(\trace_i; x, z) & \trace_j \cong_O z \mbox{ for all } j=1\ldots K\\
        0 & \mbox{otherwise}
\end{array} \right.
\end{align}
Note that $q(z, \trace_{1:K})$ can be sampled from by first sampling an index $k$ uniformly from $\{1\ldots K\}$, then executing $\prog$ to produce $\trace_k$, extracting the output choices $z := {\trace_k}|_O$, and executing $\prog$ an additional $K-1$ times using fixed output $z$ to produce $\trace_j$ for $j \in \{1\ldots K\} \setminus \{k\}$.
First note that $\pi(z, \trace_{1:K}) > 0 \implies q(z, \trace_{1:K}) > 0$ since $p(\trace_k; x, z) > 0$ implies $\trace_k \cong_O z$ and $p(\trace_k; x) > 0$.
The importance weight for the extended target distribution and the extended proposal distribution is:
\begin{equation}
\frac{\pi(z, \trace_{1:K})}{q(z, \trace_{1:K})}
= \frac{\pi(z) \prod_{i=1}^K p(\trace_i; x, z)}{\frac{1}{K} \sum_{k=1}^K p(\trace_k; x) \prod_{i \ne k} p(\trace_i; x, z)}
= \frac{\pi(z)}{\frac{1}{K} \sum_{k=1}^K \frac{p(\trace_k; x)}{p(\trace_k; x, z)} }
= \frac{\pi(z)}{\frac{1}{K} \sum_{k=1}^K p_O(\trace_k; x) }  \label{eq:importance-weight-extended}
\end{equation}
Note that invoking \textproc{simulate} in Algorithm~\ref{alg:is} is equivalent to sampling from the extended proposal distribution $q(z, \trace_{1:K})$ and that the importance weight $\pi(z)/\hat{\xi}$ in Algorithm~\ref{alg:is} is equal to Equation~(\ref{eq:importance-weight-extended}).
Therefore, Algorithm~\ref{alg:is} is a standard self-normalized importance sampler for the extended target distribution $\pi(z, \trace_{1:K})$.
By convergence of standard self-normalized importance sampling, $\hat{\mu}_N \stackrel{\mbox{\tiny a.s.}}{\to} \mathbb{E}_{(z, \trace_{1:K}) \sim \pi(\cdot)}[ f(z)] = \mathbb{E}_{z \sim \pi(\cdot)}[f(z)] = \mu$.

\subsection{Stationary distribution for Metropolis-Hastings using proposal program} \label{sec:mh-proof}
Algorithm~\ref{alg:mh} defines a transition operator that takes as input an output trace of proposal program $\prog$, denoted $x = z_{t-1}$, and produces as output another output trace, denoted $z_t$.
Note that other state besides $z_{t-1}$ (i.e. state that is not mutated by the transition operator) can be included in the input $x$ to the operator, but we write $x = z_{t-1}$ to simplify notation.
Let $\delta$ denote the Kronecker delta.
Let $\trace_{1:K}$ denote the tuple $(\trace_1, \ldots, \trace_K)$.
To show that the this transition operator admits the target distribution $\pi(z)$ as a stationary distribution, we first define an extended target distribution on tuples $(z, \zeta, k, \trace_{1:K})$ where $\zeta$ and $z$ are both output traces of $\prog$, where $k \in \{1\ldots K\}$, and where each $\trace_j$ for $j=1\dots K$ are traces of $\prog$:
\begin{equation}
    \pi(z, \zeta, k, \trace_{1:K})
    := \pi(z) \frac{1}{K} p(\trace_k; z) \delta(\zeta, {\trace_k}|_O) \prod_{i \ne k} p(\trace_i; z, \zeta)
\end{equation}
One could hypothetically sample from this distribution by first sampling $z \sim \pi(\cdot)$, then sampling $k$ uniformly from $\{1\ldots K\}$, extracting the output choices $\zeta := {\trace_k}|_O$, then executing the program on input $z$ to produce trace $\trace_k$, and executing the program $K-1$ times on input $z$ with output choices fixed to values given in $\zeta$.
We also define a proposal kernel on this extended space:
\begin{equation}
    q(z', \zeta', k', \trace_{1:K}'; z, \zeta, k, \trace_{1:K})
    := \delta(z'; \zeta) \delta(\zeta'; z) \left( \prod_{j=1}^K p(\trace_j'; \zeta, z) \right)  \frac{p_O(\trace_{k'}'; \zeta)}{\sum_{j=1}^K p_O(\trace_j'; \zeta)}
\end{equation}
This kernel can be sampled from by first executing $\prog$ on input $\zeta$, $K$ times, with output random choices fixed to values in $z$, producing traces $\trace'_{1:K}$, then sampling $k'$ from $\{1\ldots K\}$ in proportion to $p_O(\trace'_{k'}; \zeta)$, and then setting $z' := \zeta$ and $\zeta' := z$.
Consider a Metropolis-Hastings (MH) move on the extended space, using $q$ as the proposal kernel and the extended target $\pi$ as the target.
Assume that $z' = \zeta$, $\zeta' = z$, $\zeta = {\trace_k}|_O$, and $\zeta' = {\trace_{k'}|_O}$.
The MH acceptance ratio for this move is:
\begin{align}
&\frac{\pi(z', \zeta', \trace_{1:K}') q(z, \zeta, \trace_{1:K}; z', \zeta', \trace_{1:K}')}{\pi(z, \zeta, \trace_{1:K}) q(z', \zeta', \trace_{1:K}'; z, \zeta, \trace_{1:K})}\\
&= \frac{
      \pi(z') \frac{1}{K} p(\trace_k'; z') \prod_{i \ne k'} p(\trace_i'; z', \zeta')
      \left( \prod_{j=1}^K p(\trace_j; \zeta', z') \right)  \frac{p_O(\trace_{k}; \zeta')}{\sum_{j=1}^K p_O(\trace_j; \zeta')}
     }
     {
      \pi(z) \frac{1}{K} p(\trace_k; z) \prod_{i \ne k} p(\trace_i; z, \zeta)
      \left( \prod_{j=1}^K p(\trace_j'; \zeta, z) \right)  \frac{p_O(\trace_{k'}'; \zeta)}{\sum_{j=1}^K p_O(\trace_j'; \zeta)}
     }\\
&= \frac{
      \pi(z') p(\trace_k'; z')
      p(\trace_k; z, \zeta) \frac{p_O(\trace_k; \zeta')}{\sum_{j=1}^K p_O(\trace_j; \zeta')}
     }
     {
      \pi(z) p(\trace_k; z)
      p(\trace_{k'}'; \zeta, z) \frac{p_O(\trace_{k'}'; \zeta)}{\sum_{j=1}^K p_O(\trace_j'; \zeta)}
     }\\
&= \frac{
      \pi(z') p_O(\trace_{k'}; \zeta) \frac{p_O(\trace_k; \zeta')}{\sum_{j=1}^K p_O(\trace_j; \zeta')}
     }
     {
      \pi(z) p_O(\trace_k; z) \frac{p_O(\trace_{k'}'; \zeta)}{\sum_{j=1}^K p_O(\trace_j'; \zeta)}
     }\\
&= \frac{
      \pi(z') \frac{1}{K} \sum_{j=1}^K p_O(\trace_j'; \zeta)
     }
     {
      \pi(z) \frac{1}{K} \sum_{j=1}^K p_O(\trace_j; z)
     } \label{eq:mh-accept}
\end{align}
Consider a sampling process that begins with $z \sim \pi(\cdot)$ then executes Algorithm~\ref{alg:mh} on input $z_{t-1} = z$.
Observe that the invocation of \textproc{simulate} in Algorithm~\ref{alg:mh} is equivalent to sampling $(\zeta, k, \trace_{1:K})$ from $\pi(\zeta, k, \trace_{1:K} | z)$.
Therefore, after \textproc{simulate}, we have a sample from the extended target distribution: $(z, \zeta, k, \trace_{1:K}) \sim \pi(\cdot)$.
Next, note that the invocation of \textproc{assess} in Algorithm~\ref{alg:mh}, on input $z_{t-1} = z$ and $z' = \zeta$, is equivalent to sampling $(z', \zeta', k', \trace_{1:K}')$ from the extended proposal kernel on input $(z, \zeta, k, \trace_{1:K})$.
Note that $k'$ is not literally sampled in \textproc{assess} but could be without changing the behavior of \textproc{assess}.
Finally, note that Equation~(\ref{eq:mh-accept}) is the acceptance ratio used in the accept/reject step of Algorithm~\ref{alg:mh}.
Therefore, we can interpret Algorithm~\ref{alg:mh} as first extending the input sample $z$ onto the extended space by sampling from the conditional distribution $\pi(\zeta, k, \trace_{1:K} | z)$, and then performing an MH step on the extended space.
Since $z \sim \pi(\cdot)$, we have that $(z, \zeta, k, \trace_{1:K}) \sim \pi(\cdot)$.
Since the MH step on the extended space has the extended target distribution $\pi$ as a stationary distribution, we have $(z', \zeta', k', \trace_{1:K}') \sim \pi(.)$ and therefore $z' \sim \pi(\cdot)$.
Therefore $\pi(z)$ is a stationary distribution of Algorithm~\ref{alg:mh}.

\section{Offline optimization of proposal programs}

To see that $J^K(\theta) \le J(\theta)$, note that:
\begin{align}
    \E_{\trace_{1:K} \iid p(\cdot; x, \theta, z)} \left[ \hat{\xi}(\trace_{1:K}, x, \theta) \right]
    &= \E_{\trace_{1:K} \iid p(\cdot; x, \theta, z)} \left[ \frac{1}{K} \sum_{k=1}^K p_O(\trace_k; x, \theta) \right]\\
    &= \E_{\trace \sim p(\cdot; x, \theta, z)} \left[ p_O(\trace; x, \theta) \right]\\
    &= \sum_{\trace : \trace \cong_O z} \frac{p(\tau; x, \theta)}{p_O(\trace; x, \theta)} p_O(\trace; x, \theta)\\
    &= p(z; x, \theta)
\end{align}
By Jensen's inequality:
\begin{align}
    \E_{\trace_{1:K} \iid p(\cdot; x, \theta, z)} \left[ \log \hat{\xi}(\trace_{1:K}, x, \theta) \right] \le \log p(z; x, \theta)
\end{align}

\end{document}